\documentclass[final]{cvpr}

\usepackage{times}
\usepackage{epsfig}
\usepackage{graphicx}
\usepackage{amsmath}
\usepackage{amssymb}

\usepackage{epsfig}
\usepackage{graphicx}
\usepackage{amsmath}
\usepackage{amssymb}
\usepackage{booktabs}
\usepackage[table]{xcolor}
\usepackage{caption}
\usepackage{subcaption}
\usepackage{setspace}
\usepackage{bm}
\usepackage{gensymb}
\usepackage{multicol}
\usepackage{bbm}

\usepackage[utf8]{inputenc} 
\usepackage[T1]{fontenc}    
\usepackage{url}            
\usepackage{booktabs}       
\usepackage{amsfonts}       
\usepackage{nicefrac}       
\usepackage{microtype}      

\usepackage{epsfig}
\usepackage{graphicx}
\usepackage{amsmath}
\usepackage{amssymb}
\usepackage{booktabs}
\usepackage[table]{xcolor}
\usepackage{caption}
\usepackage{subcaption}
\usepackage{setspace}
\usepackage{bm}
\usepackage{gensymb}
\usepackage{multicol}
\usepackage{bbm}

\usepackage{array}

\usepackage{color,soul}

\usepackage[pagebackref=true,breaklinks=true,colorlinks,bookmarks=false]{hyperref}

\def\cvprPaperID{2034} 
\def\confYear{CVPR 2021}

\usepackage{listings}
\usepackage{xcolor}
\usepackage{textcomp}
\usepackage{color}
\definecolor{deepblue}{rgb}{0,0,0.5}
\definecolor{deepred}{rgb}{0.6,0,0}
\definecolor{deepgreen}{rgb}{0,0.5,0}

\newcommand\pythonstyle{\lstset{
language=Python,
basicstyle=\ttfamily,
otherkeywords={self},             
keywordstyle=\ttfamily\bfseries\color{deepblue},
emph={MyClass,__init__},          
emphstyle=\ttfamily\bfseries\color{deepred},    
stringstyle=\color{deepgreen},
commentstyle=\color{deepgreen},
frame=tb,                         
upquote=true,
showstringspaces=false            %
}}

\lstnewenvironment{python}[1][]
{
\pythonstyle
\lstset{#1}
}
{}

\newcommand\pythonexternal[2][]{{
\pythonstyle
\lstinputlisting[#1]{#2}}}

\newcommand\pythoninline[1]{{\pythonstyle\lstinline!#1!}}

\usepackage{graphicx}
\usepackage{comment}
\usepackage{amsmath,amssymb} 
\usepackage{color,soul}

\definecolor{partcolor}{RGB}{171, 50, 57}
\definecolor{citecol}{RGB}{49, 116, 162}
\definecolor{gray75}{gray}{0.65}
\definecolor{col2}{RGB}{224,56,189}
\definecolor{col3}{RGB}{0, 211, 234}
\definecolor{col1}{RGB}{234,180,160}

\usepackage[T1]{fontenc}
\usepackage[utf8]{inputenc}

\begin{document}

\title{Recognizing Actions in Videos from Unseen Viewpoints}

\author{AJ Piergiovanni\\
Indiana University\\
{\tt\small ajpiergi@indiana.edu}
\and
Michael S. Ryoo\\
Stony Brook University\\
{\tt\small mryoo@cs.stonybrook.edu}
}

\maketitle

\begin{abstract}
Standard methods for video recognition use large CNNs designed to capture spatio-temporal data. However, training these models requires a large amount of labeled training data, containing a wide variety of actions, scenes, settings and camera viewpoints. In this paper, we show that current convolutional neural network models are unable to recognize actions from camera viewpoints not present in their training data (i.e., unseen view action recognition). To address this, we develop approaches based on 3D representations and introduce a new geometric convolutional layer that can learn viewpoint invariant representations. Further, we introduce a new, challenging dataset for unseen view recognition and show the approaches ability to learn viewpoint invariant representations.
\end{abstract}

\section{Introduction}

Activity recognition with convolutional neural networks (CNNs) has been very successful \cite{carreira2017quo,tran2014c3d,feichtenhofer2018slowfast} when provided sufficient diverse labeled data, like Kinetics \cite{kay2017kinetics}. However, one major limitation of these CNNs is that they are unable to recognize actions/data that are outside of the training data distribution. This is most notably observed for unseen classes (objects, activities, etc.) which has been heavily studied in zero-shot and few-shot learning literature. In this work, we look at a related, but different problem of \emph{unseen viewpoint} activity recognition, where the actions are the same, but occur from different camera angles.

\begin{figure}
\centering
\includegraphics[width=0.9\linewidth]{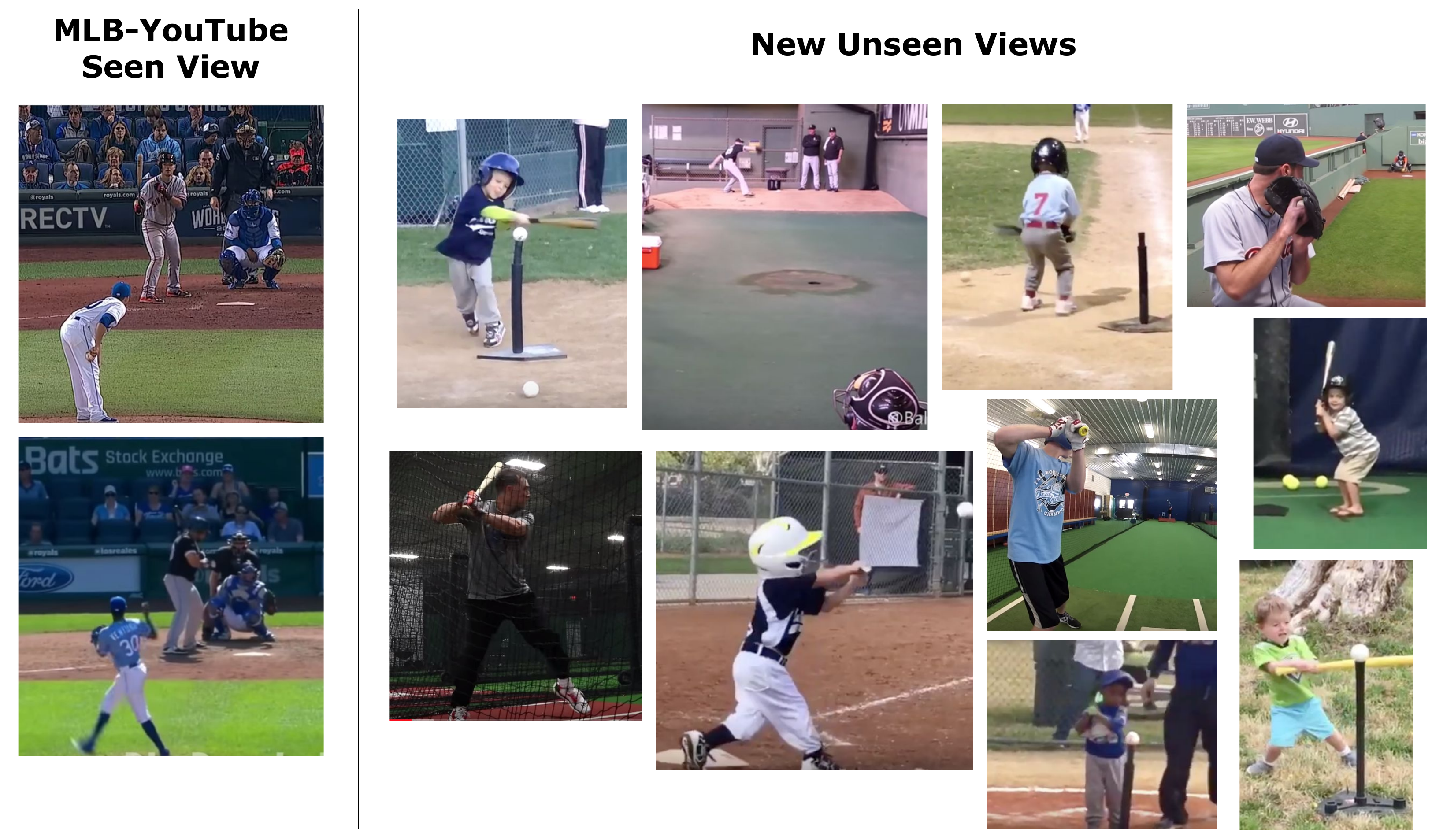}
\captionof{figure}{Examples of the seen, static broadcast camera in MLB-YouTube and examples of the new, unseen viewpoints of the same actions. This dataset is quite challenging, adding new views, people, etc.}
    \label{fig:unseen-baseball}
\end{figure}

To motivate this problem, let us consider an example. Given a labeled dataset of a person performing actions with one camera angle, we train a CNN to recognize this action. Now, suppose we have new videos to recognize, but from a different camera view. This could be as simple as a different camera placement in the environment, or an entirely different camera and setting (e.g., Fig. \ref{fig:unseen-baseball}). In this case, a trained CNN, in general, fails to recognize the action. As a simple experiment, we use the Human3.6M dataset \cite{h36m}, which contains videos of a person performing an action from 4 different camera angles. As shown in Table \ref{tab:h36m-base}, when training on one view and testing on another, the model is unable to recognize the action. However, humans are able to recognize these actions regardless of viewpoint and studies have found that this is likely because humans build invariant representations of actions in their minds \cite{tacchetti2017invariant} .

Further, this problem frequently occurs in real data (e.g., YouTube videos). Existing smaller datasets such as Toyota SmartHome \cite{toyotasmarthome}, Charades-Ego \cite{sigurdsson2018charadesego}, NTU \cite{shahroudy2016ntu} and others all provide videos in multiple viewpoints to study this effect. Large video datasets like Kinetics \cite{kay2017kinetics} naturally contain many views, however, there is no annotation of the view and each video only provides a single view. Other datasets like MLB-YouTue \cite{mlbyoutube2018} only contains the single broadcast camera view baseball games. As collecting video data is already challenging, designing CNNs that generalize to unseen viewpoints is critical, especially for applications where diverse view data is limited or unavailable. It would be practically impossible to build datasets for many desirable settings that enumerate all possible (or sufficiently large number of) viewpoints to fully model activities.

\begin{table}
\centering
    \begin{tabular}{l|cc}
    \toprule
    Method & Seen & Unseen\\
    \midrule
    Random & 9.1\% & 9.1\% \\
    2D ResNet-50 & 86.4\% & 9.1\%  \\
    3D ResNet-50 & 100\% & 9.1\% \\
    3D ResNet-50 + Kinetics & 100\% & 38.2\%\\
    Ground Truth 3D Pose & 100\% & 100\% \\
    \bottomrule
    \end{tabular}
    \caption{Experiments on Human3.6M with unseen viewpoints. Standard CNNs are unable to recognize actions with different viewpoints, however, using global 3D pose allows the models to recognize the actions.}
    \label{tab:h36m-base}
\end{table}

There are many potential ways to address this problem. One hypothesis is that by training on a large-scale video datasets, such as Kinetics, the model could implicitly learn multi-view representations of actions. However, as shown in Table \ref{tab:h36m-base}, we empirically find that while it improves performance, it is still lacking. A second hypothesis is that by using 3D human pose information, we can recognize actions in a global representation space, unconstrained by camera views. A key drawback to this approach is estimating 3D pose from video itself is a challenging problem, especially when multiple people are present. It further requires estimating camera pose in order to build a `world'/global camera invariant 3D representation. Further, it is unclear what the right representation of 3D pose is (e.g., coordinates of joints, limbs, motion difference of joints between frames, etc.).

Building on this hypothesis and observation, we present and evaluate several approaches for recognizing actions in unseen viewpoints. The basic approach relies on estimating 3D pose directly from the videos, then explores using different representations of it for recognition. Since directly estimating accurate real-world 3D pose is often difficult, we also present an approach of learning \emph{latent} 3D representations of an action and its multi-view 2D projections. This is done by imposing the latent action representations to follow 3D geometric transformations and projections, in addition to minimizing the action classification loss. We learn such view invariant action representations \emph{without} any 3D or view ground truth labels. 

We also introduce a challenging dataset building on the MLB-YouTube dataset \cite{mlbyoutube2018}. The MLB-YouTube dataset contains actions from a single camera and these actions are all in the same environment (e.g., a professional baseball stadium). Our extended dataset contains evaluation samples of the same actions, but from many different viewpoints and a variety of different settings: batting cages, little league (children's baseball games), high school games, etc. These use different camera (e.g., cell phones), in very different environments. The goal is to learn a representation from the single view dataset that generalizes to these challenging, unseen viewpoints. Examples are shown in Fig. \ref{fig:unseen-baseball}.

To summarize, the contributions of this paper are:
\begin{itemize}  \setlength\itemsep{0em}
    \item A computationally efficient, geometric-based layer and learning to learn view invariant representation.
    \item Thorough evaluation of multiple approaches to unseen viewpoint action recognition.
    \item A challenging new dataset for unseen viewpoint recognition in unseen environments.
\end{itemize}

\section{Background - 3D Geometric Camera Model}
First we briefly review the standard 3D geometric camera model used in computer vision, which we build on in this work. We begin with a standard pinhole camera model. Given the pixel coordinates $p$, the 3D world coordinates $p_w$ are represented as 
\begin{equation}
    p = \bm{K}~[~\bm{R}~|~\bm{t}~]~[~p_w~|~\bm{1}~]^T
\end{equation}
where $\bm{K}$ is the $3\times 4$ camera projection matrix (intrinsic camera matrix) mapping a 3D point into a 2D camera view. $\bm{R}$ and $\bm{t}$ are the camera rotation ($3\times 3)$ and translation ($1\times 3$) in the world space (extrinsic camera matrix) that transform the points between different 3D camera views. $|$ is the matrix concatenation operator. In many cases in computer graphics and computer vision, it is assumed that $\bm{K}$, $\bm{R}$ and $\bm{t}$ are known. These matrices can be used to compute the inverse as $[~\bm{R}^T~|~\bm{R}^T\bm{t}~]$ (the inverse of a rotation matrix is its transpose). Thus, given a 3D coordinates $p$, the view-invariant world coordinates can be computed as $p_w = p\cdot~[~\bm{R}^T~|~\bm{R}^T\bm{t}~]$.

Importantly, points in the 3D world coordinate system are, by design, viewpoint invariant. However, in many settings, including activity recognition, the camera matrices are unknown. In some cases the intrinsic camera matrix $\bm{K}$ might be known (e.g., when the camera has been previously calibrated), but the extrinsic matrices as well as the definition of the world coordinate system are not. The core of this paper is exploring methods to learn and represent these.

\section{Basics - Using 3D Pose}
\label{sec:a1}

\begin{figure}
    \centering
    \includegraphics[width=0.75\linewidth]{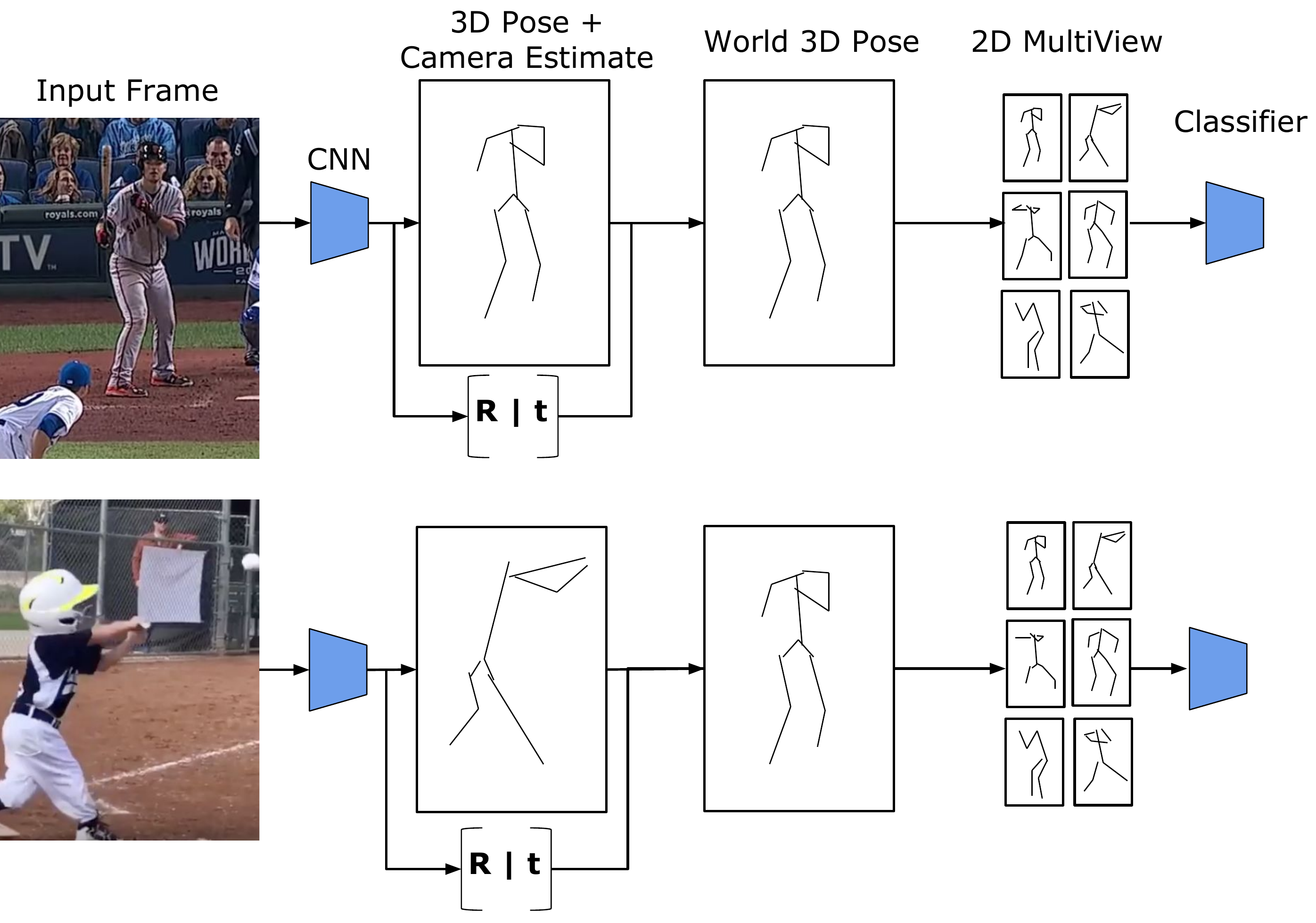}
    \caption{Overview of the process to learn global 3D pose and 2d multi-view projections of it for classifying unseen viewpoints.}
    \label{fig:model-overview}
\end{figure}

We first design and investigate a straight forward approach of using 3D human pose estimation and its projection for action classification (Fig. \ref{fig:model-overview}). Many works have explored estimating 3D human pose from videos \cite{pavllo20193dpose,kanazawa2019learning,li20143dpose,li2015max3dpose,moreno20173dpose}, even multi-person 3D pose \cite{multiperson3dpose}. We begin by using PoseNet \cite{multiperson3dpose} to estimate 3D coordinates. PoseNet provides 3D coordinates in camera space, so directly using the coordinates will not yield viewpoint-invariant recognition.  For this, the 3D coordinates need to be transformed into world-space.

However, estimating the extrinsic matrix from a single, random image is a challenging problem. We use CalibNet \cite{iyer2018calibnet} to obtain estimations of $\bm{R}$ and $\bm{t}$. This approach is quite limited by the accuracy of CalibNet, if it provides a poor estimation, the rest of the network will fail. Since there is limited camera calibration training data, we observe than for in-the-wild videos, CalibNet often gives inaccurate results and does not generalize well.

\subsection{Recognition with 3D representation}
\label{sec:multi-view}
Given the estimated 3D pose in world space, these values can be directly used as input to the model. However, directly using 3D coordinates may not be the best feature, as scale changes (e.g., person size), speed of which the action occurs, etc. will all impact the features. Previous works have studied representations of 3D pose for skeleton action recognition \cite{du2015hierarchicalskeleton,yan2018spatialskeleton,ke2017newskeleton,li2017skeleton}, such as joint angles \cite{ohn2013joint} showing the difficulty of this task.

Instead of directly using the 3D pose as input to a classification model, multi-view 2D projections of it could be used. Research focusing on designing strong CNNs for understanding 2D image input has been one of the mainstream areas, and it often is more advantageous to use 2D image inputs rather than 3D. Further, by using multiple views, the model can see the input from different angles, instead of a single one. To do this, we assume we have an intrinsic camera matrix $K$ that projects the 3D coordinates into 2D. We follow the standard pinhole camera model and learn a camera rotation and projection to generate multiple 2D views. Inspired by previous works, like Potion \cite{choutas2018potion}, we take the 2D projections of pose and render skeletons capturing the motion. These images are used as input to the model for activity classification. An overview of this approach is shown in Fig. \ref{fig:model-overview}.

\section{Network for Latent 3D Representations}
\label{sec:a2}
The previous approach is an engineered combination of existing CNNs (pose, camera estimation and action recognition) and relies on multiple components correctly functioning. If any of these networks fail or gives slightly incorrect results, the rest of the model will fail. However, we draw inspiration from the geometric based approach, and design geometric CNN layers to learn and replicate similar transformations. To do this, we begin by learning a representation that contains and uses both 3D information and the extrinsic information. We then combine this information to get a 3D world representations of the actions and provide a CNN architecture. We introduce a loss function terms to learn such 3D view-invariant representations from a dataset with (unlabeled) views.

\subsection{Neural Projection Layer (NPL): Building a Latent 3D World Representation}
First, we take a feature map $F$ which is a $W\times H\times (C+3)$ tensor, where $C$ is the channels in the feature map plus the CNN estimated 3D camera space coordinate. Formally, $F_{x',y'} = [p_{x',y'}, f_{x','y}]$, where $p_{x',y'} = [x,y,z]$ at location $x',y'$ in the feature map and $f_{x',y'}$ is the $C$-dimensional feature at the location.

Next, we use a fully-connected layer to estimate $\bm{R}$ and $\bm{t}$ (rotation and translation) from each video. These are used to transform the video camera view into the world coordinates as $p^w_{x',y'} = p_{x',y'}\cdot[~\bm{R}^T~|~\bm{R}^T\bm{t}~]$ for each $p$ in the feature map. This gives the 3D world coordinates, i.e., camera invariant coordinate, of each point $p$.  Note that the 3D world coordinate system is the same for all videos, thus $\bm{R}$ is different for each video, depending on the camera viewpoint. $\bm{R}$ plays the role of aligning features (e.g., humans) in different scenes, so that the losses are minimized. This allows the model to learn to map each video into the same global coordinates. 

The world 3D representation is then created as:
\begin{equation}
  F^W_{x,y,z} = \sum_{i=0,j=0}^{W,H} \mathbbm{1}(p^w_{i,j} = [x,y,z]) F_{i,j}
\end{equation}
where $\mathbbm{1}(p^w_{i,j} = [x,y,z])$ is the indicator function for when $p^w$ matches location $x,y,z$. 

That is, we can create a new feature map $F^W$ which has a shape of $W\times H\times Z$ (in practice, $W=H=Z$, for example, 64). Given one of the points $p_{x',y'}$ and its associated feature vector $f_{x',y'}$ from the original feature map, we compute the `world coordinate' of that point $x,y,z = p^w_{x',y'}$ and then set $F^W_{x,y,z} = F^W_{p^w_{x',y'}} = f_{x',y'}$. I.e., we set the values in the feature map based on their location in the world coordinate reference. This transforms the original latent 3D representation into the rotation and translation invariant 3D world representation, resulting in a representation that is viewpoint/camera invariant.

However, since $x,y,z$ are integers and $p^w_{i,j}$ is likely not an integer and we want to implement this as a differentiable function to be learned with gradient descent, we slightly modify this to:
\begin{equation}
\label{eq:3drep}
\begin{split}
 F^W_{c,x,y,z} = \sum_{i=0,j=0}^{W,H} &(1-|x - p^w_{i,j}[x]|)\\
 &(1-|y - p^w_{i,j}[y]|)(1-|z - p^w_{i,j}[z]|) F_{c,i,j}
\end{split}
\end{equation}
This is similar to Eq. 5 introduced in the spatial transformer network \cite{jaderberg2015spatial}. In the implementation, we first set $p_w = \tanh{p^w}$ to ensure its values fit in the feature map space, similar to the transformer network.

Once we obtain $F_W$, we use it directly as input to the remaining CNN for classification. We note that $F_W$ is a 4-dimensional tensor (channels followed by 3D coordinates), so we use 3D convolution on top of this representation.

\begin{figure}
    \centering
    \includegraphics[width=0.75\linewidth]{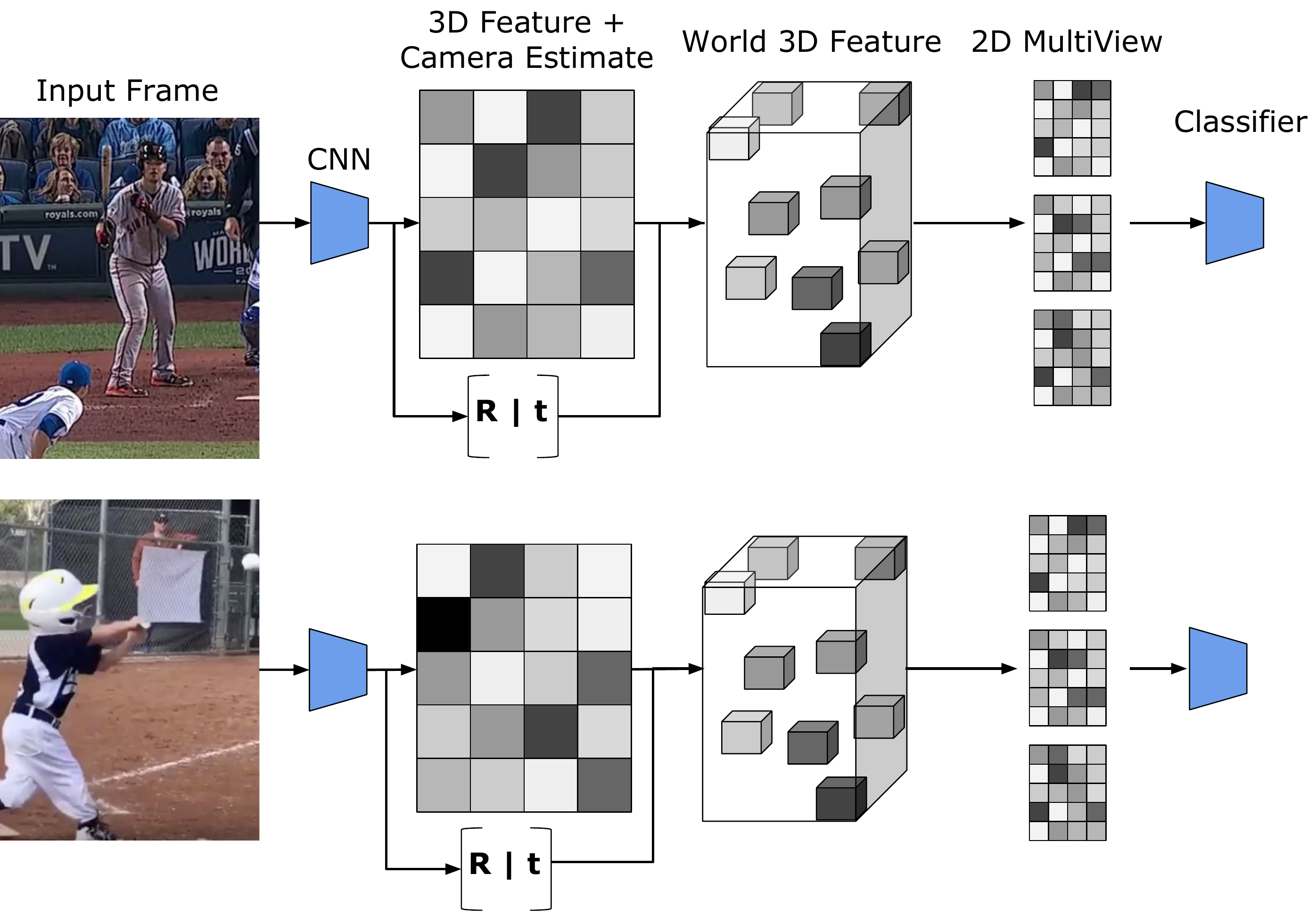}
    \caption{Learning latent 3D representation. A CNN produces a 3D feature: each location has a feature value and $x,y,z$ location. The network also produces camera parameters, allowing the construction of a viewpoint invariant 3D feature. Then multiple cameras are learned, allowing the creation of multi-view 2D projections of the features. These are stacked on the channel axis and used for classification. The world 3D feature and 2D MultiView features are learned to be identical for the same action.}
    \label{fig:latent-model}
\end{figure}

\subsubsection{Multi-view 2D Projections}
Instead of directly working with a 3D feature map, we can follow the ideas from Section \ref{sec:multi-view} and generate multiple 2D projections of the features. This is shown in Fig. \ref{fig:latent-model}.

Assuming we have a camera matrix $\bm{K}$, which we represent as 
\begin{equation}
    \bm{K} = \bm{R} \left ( 
                \begin{array}{ c c c}
                s_x & 0   & x_0 \\
                 0  & s_y & y_0 \\
                 0  & 0   & 1 \\
                \end{array}
            \right )
\end{equation}
where $s_x,s_y$ are the focal lengths and $x_0,y_0$ are the offsets, and $\bm{R}$ is the $3\times 3$ camera rotation matrix. Note $\bm{R}$ can be represented with 3 parameters: yaw, pitch and roll which generate the full $3\times 3$ rotation matrix. This process uses the same components and projections as in Section 4.1, but instead of estimating $\bm{R}$ using a layer for each video, here these are learned parameters of the model. This allows the model to generate the same 2D views from the view-invariant world 3D space. We can then model a 2D projection of the 3D points as:

\begin{equation}
\label{eq:mvrep}
  F^W_{c,x,y} = \sum_{i=0,j=0}^{W,H} (1-|x - \bm{K}p^w_{i,j}[x]|)(1-|y - \bm{K}p^w_{i,j}[y]|) F_{c,i,j}
\end{equation}

As these operations are differentiable, we can learn all these parameters with gradient descent. This allows the model to learn the optimal arrangement of cameras to capture views from the latent 3D representations for action recognition. Further, by increasing the number of cameras ($N$), we can learn multiple 2D projections of the representations, which can be stacked on the channel axis for recognition. In the next section, we describe the training loss that enables learning of the 3D representation without any 3D or camera calibration ground truth data. 

We note that in this setting, some views will have occluded objects/features. An assumption of the approach is that using multiple cameras will naturally capture different viewpoints that will minimize the effects of occlusion. Another approach could be to use tomographies of each view to remove the effect of occlusions, which we leave as future work.

\subsection{Recognition with 3D Representation}
\begin{figure}
    \centering
    \includegraphics[width=0.8\linewidth]{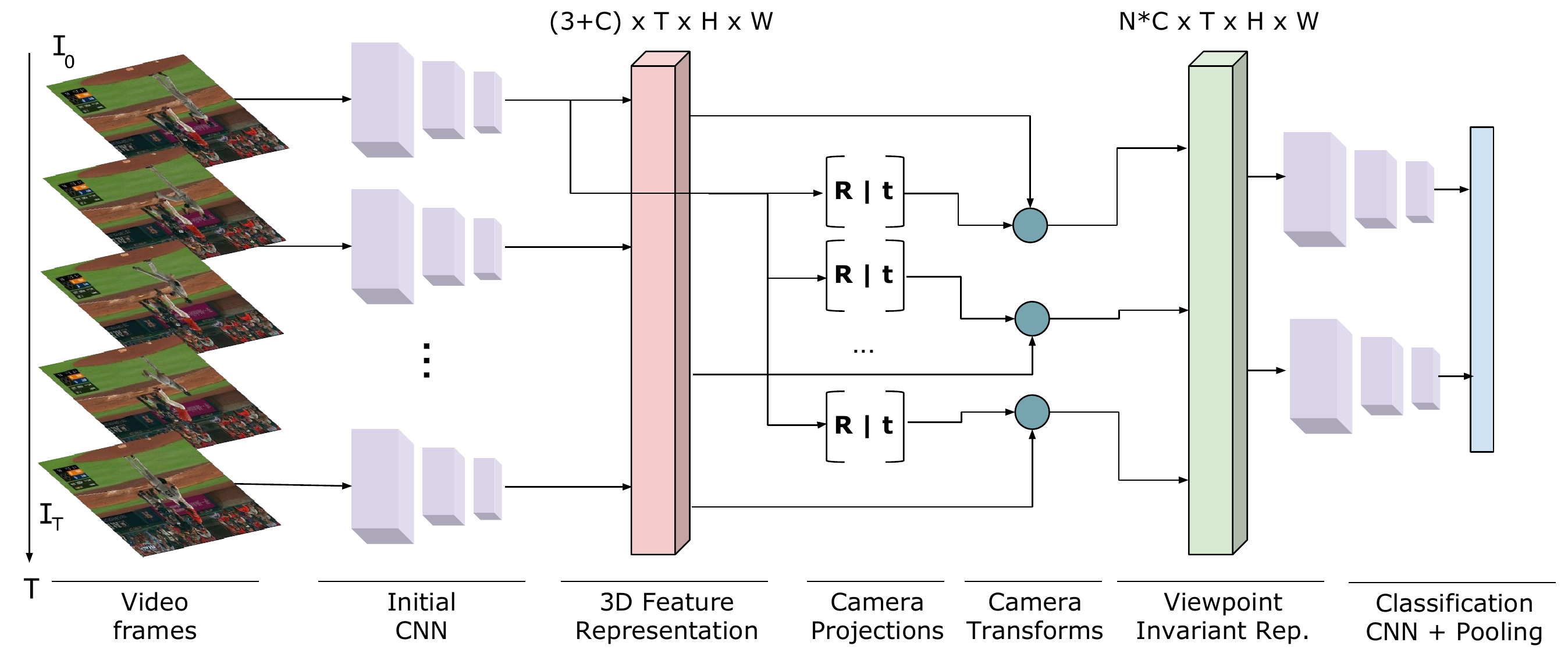}
    \caption{Illustration of the full viewpoint invariant recognition model.}
    \label{fig:model}
\end{figure}

The full model, as illustrated in Fig. \ref{fig:model}, begins by applying several conv. layers to a video input. At some point in the network (we experimentally evaluate where), we generate the 3D feature map and apply the geometric transformations described above. We then use either 2D or 3D conv. layers followed by a fully connected layer to classify the video clip.

\section{Learning Latent 3D Representations}
The key challenge with this approach is learning the camera matrices that generate view-invariant representations. We propose an approach to learn this view-invariant representation from a dataset with action videos (e.g., Kinetics \cite{kay2017kinetics}). Importantly, we do not assume any ground truth viewpoint or 3D data is provided, but only that the videos naturally contain multiple views. Unlike the first approach where we first learn 3D human pose estimation and extrinsic camera calibration from specific datasets, this approach only requires action-labeled data, available in many large-scale public datasets.

Given two videos $V$ and $U$ of the same action, we compute their 3D representations $F_W(V)$ and $F_W(U)$ and use a loss to make their representations the same:
\begin{equation}
\label{eq:3dloss}
    \textrm{3d\_loss}(V,U) = ||F_W(V) - F_W(U)||_F
\end{equation}

Intuitively, this loss term makes it so that two videos likely with different camera views have the same 3D representation after the projections. This encourages the representations from different viewpoints of the same action result in the same 3D representation. 

For multi-view 2D projection, we apply that loss on each 2D view, as well as adding a regularizing term to make each camera different:
\begin{equation}
\label{eq:camreg}
    cam\_reg(c_1,c_2) = \max (-||c_1-c_2||_F, \alpha)
\end{equation}
where $c_1$, $c_2$ are camera matrices and $\alpha$ is the margin, or desired max difference. We note that this constraint forces the cameras to be different, but does not ensure that they are facing the scene. However, we observed that during training, the cameras converged to views that were facing the scene, as shown in Fig. \ref{fig:cameras}.

\subsection{Training Loss Function}
Our final loss is a combination of these terms plus a standard classification loss. Given the set of cameras in the model $\mathcal{C}$, let $V$ be a video and $l$ be the binary vector indicating the class label of a video. Let $M(V)$ be the application of the network to video $V$ giving the predictions for each class (e.g., a $c$-dimensional vector where $c$ is the number of classes). In particular, $M(V)$ is some video CNN that produces the classification. More details are shown in Fig. \ref{fig:model} and specific architecture details are in the appendix. Given the set of training videos and labels, $\mathcal{V}$, The final loss function is:
\begin{equation}
\begin{split}
    \mathcal{L}(\mathcal{V}) &= \sum_{(V,l)\in\mathcal{V}} \left( -\sum_i^c l_i \log  M(V)_i \right)\\
    + &\lambda_1 \left(\sum_{(U,k)\in\mathcal{V}}
    \begin{cases} 
    \textrm{3d\_loss}(V,U) & l = k\\
    0 & otherwise
    \end{cases}
     \right)\\
    &+ \lambda_2 \left( \sum_{c_1\neq c_2\in \mathcal{C}} cam\_reg(c_1,c_2) \right)
\end{split}
\end{equation}

The combination of the geometric structure imposed by the NPL and the components of the loss function encourages the model to learn viewpoint invariant representations. This formulation enables learning 3D representations with only activity-level labels and the geometric constraints.

\section{Experiments}
We conduct various experiments to understand the various components of the approach on multiple datasets. The model was implemented in PyTorch (code in appendix) and pretrained on Kinetics-400 \cite{kay2017kinetics}. We used Kinetics to pretrain as it is large and naturally has many viewpoints, allowing for the evaluation of this approach. We then use the network to extract features and train a small two-layer network on each specific dataset. Training was done for 25 epochs with the learning rate set to 0.01. For learning the rotation matrices, we learn 3 parameters: yaw, pitch and roll, which we convert to the rotation matrix.

\paragraph{Datasets:} We evaluate this approach on three datasets containing multi-view data. On Human3.6M, we train the model on one camera view for one subject and test on one of the other views. The model is trained to recognize 11 different activities. We perform this experiment for 9 subjects and 2 different seen/unseen view combinations and report the average over each setting.

For the unseen MLB (baseball) videos, we train on the broadcast camera videos (original MLB-YouTube dataset \cite{mlbyoutube2018} and test on the new, unseen views. We newly created this dataset of unseen views for testing only. It consists of 500 videos from YouTube for testing of 4 different baseball actions (swing, hit, pitch, bunt). This data is quite challenging as it has drastically different viewpoints, people, backgrounds, activity speeds, etc. Examples of these views are shown in Fig. \ref{fig:unseen-baseball}.

For Toyota SmartHome (TSH), we follow the $CV_1$ protocol \cite{toyotasmarthome} where the model is trained using camera 1 and tested on camera 2. We also report results on the $CV_2$ protocol where it is evaluated on camera 2 but trained on multiple cameras. This dataset has 16.1k videos taken from 7 different camera viewpoints.  It contains 31 classes of human daily activities in real-world environments. Similarly, we also compare on NTU-RGB-D \cite{shahroudy2016ntu} following the standard settings.

\begin{table*}[]
    \centering
    \caption{Comparison of the approaches. While pre-training on Kinetics improves results, the use of the geometric NPL is quite beneficial.}
    \label{tab:results-3d}
    \begin{tabular}{l|cc|cc|cc}
    \toprule
    Method & \multicolumn{2}{c|}{H3.6M} & \multicolumn{2}{c|}{MLB} & \multicolumn{2}{c}{TSH} \\
          & Seen & Unseen & Seen & Unseen & Seen & Unseen \\
    \midrule
    Random & 9.1\% & 9.1\% & 25\% & 25\% & 5.2\% & 5.2\% \\
    ResNet-50 & 86.4\% & 9.1\% & 49.4\% & 27.3\%  & 34.6\% & 33.7\% \\
    ResNet-50 + Kinetics & \textbf{100\%} & 38.2\% & 55.6\% & 30.2\%  &  49.8\% & 34.2\% \\
    \midrule
    \multicolumn{5}{l}{\textbf{3D Pose Based (Section \ref{sec:a1})}} \\
    \midrule
    Ground Truth 3D Pose & \textbf{100\%} & \textbf{100\%} & - & - & 19.6\% & 14.5\% \\
    Estimated 3D Pose & 97.8\% & 78.6\% & 36.5\% & 33.6\% & 17.9\% & 11.6\% \\
    MultiView 2D Pose & 98.3\% & 81.3\% & 37.6\% & 34.6\% & 18.4\% & 12.2\% \\
    \midrule
    \multicolumn{5}{l}{\textbf{Latent 3D Representation (Section \ref{sec:a2})}} \\
    \midrule
    NPL & 99.3\% & 84.4\% & 52.3\% & 34.5\% & 51.2\% & 34.7\% \\
    NPL + Multi-view projections & 99.7\% & 87.5\% & \textbf{58.9\%} & \textbf{42.7\%} & \textbf{54.5\%} & \textbf{39.6\%} \\
    \bottomrule
    \end{tabular}
\end{table*}

In Table \ref{tab:results-3d}, we compare the different proposed approaches. We find that for some datasets, like Human3.6M, using 3D pose (Section \ref{sec:a1}) is a very good feature, as this dataset contains actions focused on human motion. However, on TSH, which has many object-dependent actions, using pose degrades performance. For example, actions such as `pick up cup' and `pick up plate', the pose motion is nearly identical for both
of those, but the object is different. Using pose gives little
indication which object is being used, thus when only using
pose to recognize the action, the model cannot distinguish.

On the MLB dataset, using pose harms performance on the seen viewpoint, but slightly improves performance on the unseen views. In both cases, it is only slightly better than random. This is likely due to noisy data with many people present, and thus the 3D pose estimation is not accurate enough. When using the learned, latent 3D representation (Section \ref{sec:a2}), we find in all cases the performance on the unseen views (and often seen views) improves, showing the benefit of the proposed approach. We note that the latent 3D representation generalizes quite well to challenging data with very different views and backgrounds (e.g., MLB data) because it is trained on large-scale video data, the model is able to learn more general projections.

We also find that training from scratch (ResNet-50) gives very poor performance on the unseen views. Somewhat surprising, pre-training on Kinetics, which has many different views of people performing actions, does slightly improve performance on unseen views, but still remains low, especially in the MLB and Toyota SmartHome datasets. This suggests that standard video CNNs are not learning 3D rotation invariant representations, even when given training data from many views, further showing the benefit of learning 3D view invariant representations.

\subsection{Ablation Experiments}
To better understand the effect of the NPL, we conduct a set of experiments to analyze each component's impact. The results are shown in Table \ref{tab:effects}. Adding the multi-view projection without any of the loss constraints slightly reduces performance. Adding the 3D loss enforces the geometric constraints to learn the viewpoint invariant representation, without this, the model struggles to learn the representation. Further adding the the camera regularization loss improves performance. Based on this observation, we also try a baseline with representation matching (RepMatch) where we apply the 3D loss (Eq. \ref{eq:3dloss}) to feature maps from a ResNet-50 without using any of the geometric layers. The findings shown no real benefit of RepMatch over standard pre-training, while the proposed geometric approach shows meaningful benefit.

\begin{table}
    \centering
    \small
    \caption{Comparison of the components of the approach. All models are based on a ResNet-50 and pretrained on Kinetics-400.}
    \label{tab:effects}
    \begin{tabular}{c|cc|cc}
    \toprule
    Method & \multicolumn{2}{c|}{MLB} & \multicolumn{2}{c}{TSH} \\
    & Seen & Unseen & Seen & Unseen \\
    \midrule
    ResNet-50 Baseline & 55.6 & 30.2 & 49.8 & 34.2 \\
    RepMatch & 55.7 & 30.3 & 48.7 & 33.8 \\
     + MultiView Proj. (MVP)    & 52.7 & 28.5 & 47.8 & 33.7 \\
     + MVP + 3D loss (Eq. \ref{eq:3dloss}) & 57.9 & 35.5 & 52.3 & 37.8 \\
     + MVP + cam reg (Eq. \ref{eq:camreg}) & 54.7 & 30.8 & 50.7 & 34.4 \\
     + MVP + 3D loss + cam reg & 58.9 & 42.7 & 54.5 & 39.6 \\
    \bottomrule
    \end{tabular}
\end{table}

\begin{figure}
    \centering
    \includegraphics{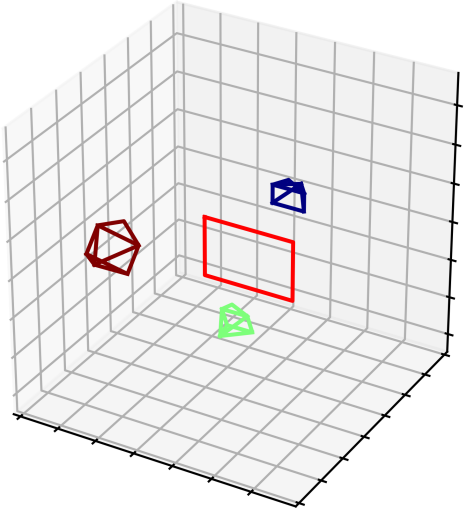}
    \caption{Visualization of the learned 2D multi-view cameras (Sec. 4.1.1). The red square represents the origin of the world coordinate system, the cameras are drawn using their intrinsic and extrinsic matrices using the matlab PlotCamera function.}
    \label{fig:cameras}
\end{figure}

We also study the effect of the number of cameras in Fig \ref{tab:numcamera}, finding that using just 1 camera projection is very helpful while more than 4 no longer improves performance. This intuitively makes sense, as a single camera will already result in a viewpoint invariant representation, and the amount of new data introduced with additional cameras decreases as more are added. In Fig. \ref{tab:where}, we compare the effect of placing the geometric layer at different locations in the network. Overall, the performance is fairly stable regardless of where it is added. In Fig. \ref{fig:cameras} we visualize the learned cameras from Sec 4.1.1. In the figure, the red rectangle represents the world 3D representations space which contains the CNN features $F_W$ (Eq. \ref{eq:3drep}). The brown, blue and green markers indicate the different learned camera matrices that capture different 2D views of the space (Eq. \ref{eq:mvrep}).


\begin{figure*}
    \centering
    \begin{minipage}{0.48\linewidth}
    \centering
        \includegraphics[width=0.9\linewidth]{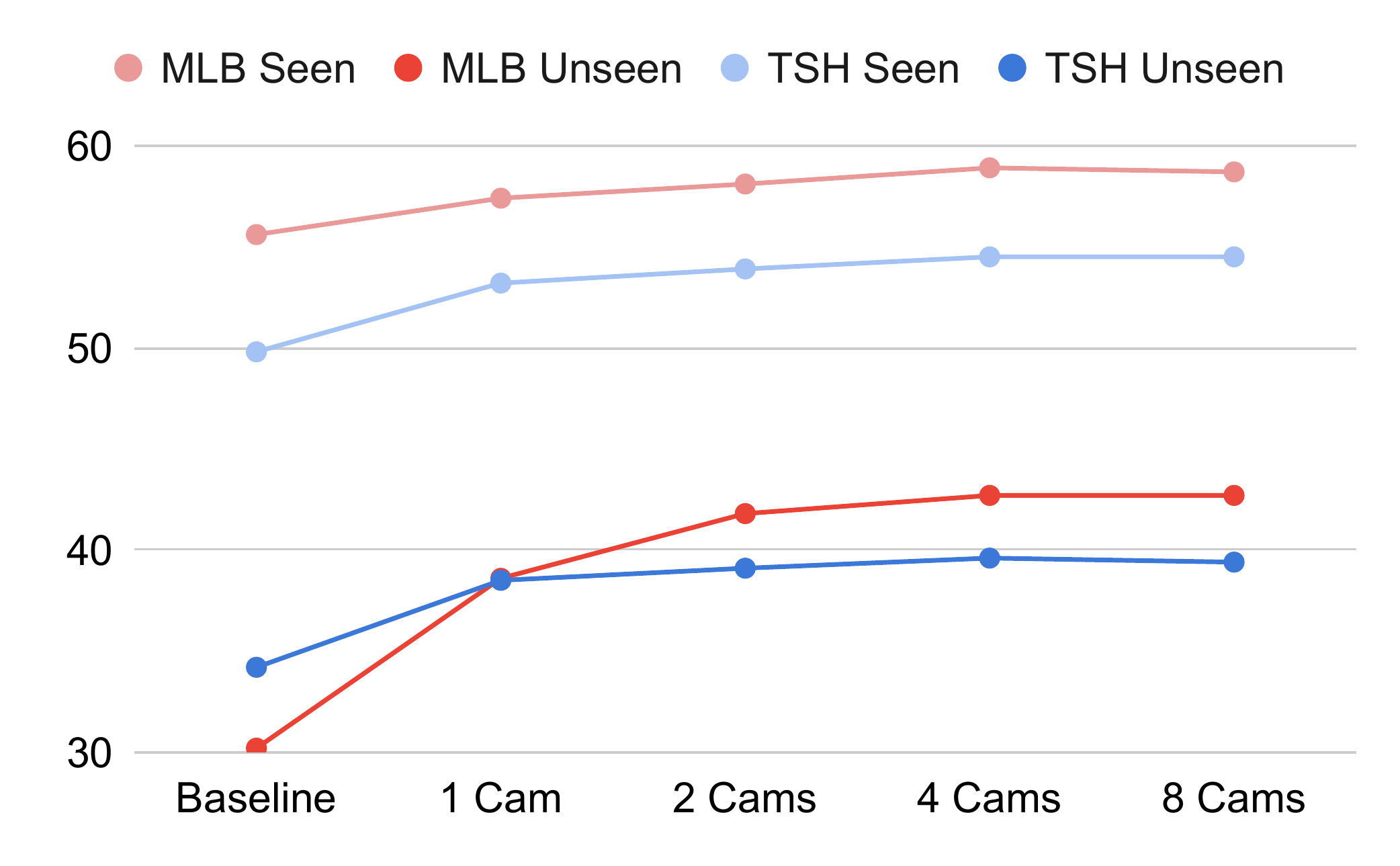}
        \caption{How many cameras to use.}
        \label{tab:numcamera}
        \end{minipage}\hfill%
\begin{minipage}{0.48\linewidth}
    \centering
            \includegraphics[width=0.9\linewidth]{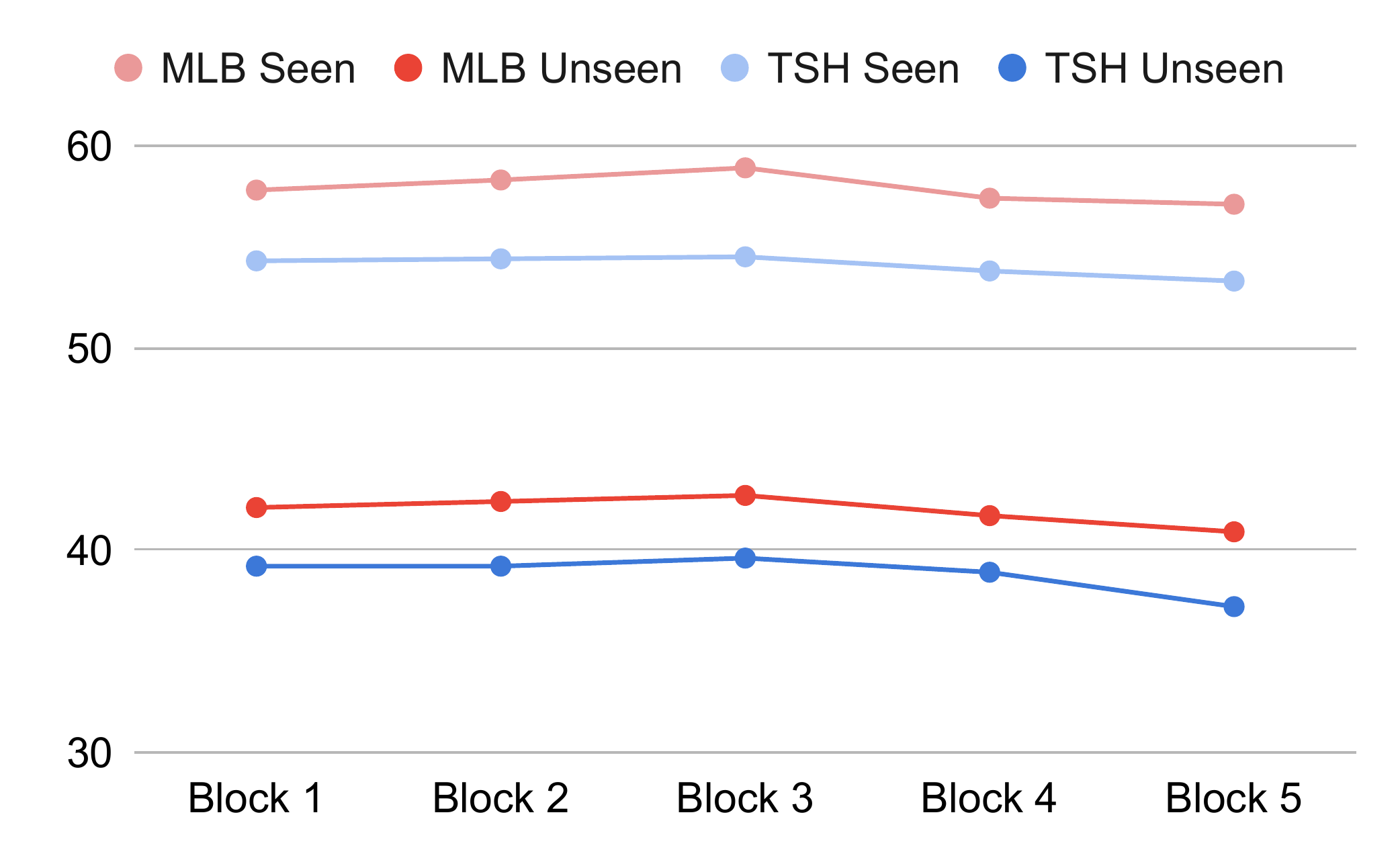}
        \caption{Where in network to add layer.}
    \label{tab:where}
\end{minipage}
\end{figure*}

\subsection{Comparison to other approaches}
We compare the proposed approach to other approaches for unseen viewpoint activity recognition. The results are shown in Table \ref{tab:approaches}, showing that the proposed approach outperforms the existing ones.  Importantly, the added runtime of our approach is small, processing a 3 second video clip in 120ms (ours) vs. 105ms (baseline ResNet-50), enabling practical use.

\begin{table}[]
    \centering
    \small
    \caption{Comparison to other approaches on Toyota SmartHome (TSH) and the Unseen MLB Views and NTU-RGB+D.}
    \label{tab:approaches}
    \begin{tabular}{c|cc|c|c}
    \toprule
    Method & \multicolumn{2}{c|}{TSH} & MLB & NTU \\
    & $CV_1$ & $CV_2$ & Unseen & $CV$ \\
    \midrule
    IDT \cite{wang2011action} & 20.9 & 23.7 & 27.3 & - \\
    Pose LSTM \cite{toyotasmarthome} & 13.4 & 17.2 & - & - \\
    I3D \cite{carreira2017quo} & 34.9 & 45.1 & 30.1 & -\\
    STA \cite{toyotasmarthome} & 35.2 & 50.3 & -  & 94.6\\
    PEM \cite{pem} & - & -  & - & \textbf{95.2} \\
    Ours & \textbf{39.6} & \textbf{54.6} & \textbf{42.7} & 93.7 \\
    \bottomrule
    \end{tabular}
\end{table}

\section{Related Works}
\paragraph{Representation Invariant Networks:} Many  works have studied representation invariant networks. The spatial transformer network \cite{jaderberg2015spatial} and Equivarient CNNs \cite{esteves2019equivariant} introduced an operation to make CNNs invariant to 2D translation, scale, rotation and more generic warping. Spherical CNNs \cite{esteves2018sphere,cohen2018spherical} took advantage of spherical representation that are invariant to 3D rotation transformations of objects in the camera view. Our approach shares some similar ideas and motivations to spherical CNNs, i.e., trying to learn a rotation invariant representation. But differs in the goal of learning object rotation invariance in spherical CNNs vs. world space representations in this work. Another difference is in the design of the representation: spherical CNNs rely on convolutions in the spherical harmonic domain, while the proposed approach uses traditional geometric computer vision to learn a representation. Other works like geometry-aware RNNs \cite{cheng2018geometry} propose the related idea of `unprojection' for learning 3D representations by utilizing ground truth 3D data.

\paragraph{View Invariant Action Recognition:} Many works have studied view invariance in action recognition \cite{viewinvar1,bashir2006trajsim,viewinvar3,viewinvar4,shen2009view,iosifidis2012view}. Several works have studied using multiple views during training to learn view invariant representations \cite{wu2013crossviewrec} or `hallucinating' features (e.g., HOG) in different viewpoints to recognize actions in unseen views \cite{chen2014inferringunseenview}. Several works explored using cross-view similarity to recognize actions in various viewpoints \cite{junejo2008crossviewsim} or trajectory curvature \cite{bashir2006trajsim}. Other works \cite{novelview2016} explored using 3D Pose for recognition, but as described, pose has limitations when interacting with objects.

\paragraph{3D Representations:} Other works have designed CNNs to specifically model 3D shapes, such as RotationNet \cite{kanezaki2018rotationnet} and others (e.g., ShapeNet, PointNet, etc.) \cite{tatarchenko2016multi3dview,chang2015shapenet,qi2017pointnet}. However, these works focused on learning 3D models, rather than applications to noisy, real videos in various environments where no 3D data is directly given.

Works such as SynSin \cite{wiles2020synsin} propose similar ideas of using geometric projections on CNN representations, showing the promise of such ideas.

\section{Conclusions}
We presented a new geometric based layer to learn 3D viewpoint invariant representations within CNNs. We also introduced a new, challenging dataset to evaluate camera view invariance. We experimentally showed the benefit of the proposed layer on multiple datasets.

\section*{Acknowledgement}
This work was supported in part by the National Science Foundation (IIS-2104404 and CNS-2104416).

{\small
\bibliographystyle{ieee_fullname}
\bibliography{egbib}
}

\clearpage
\newpage

\appendix

\section{Architecture Details}
Our base model used a standard (2+1)D ResNet-50 \cite{tran2018closer}. The camera transform is inserted into the network usually after the 3rd block (in the main paper we compared all locations). Usually this network used 256 channels for the representation and we used 3 cameras (i.e., 3 different 2D projections). The total number of parameters of the 3 main models is summarized in Table \ref{tab:params}. Our layer adds only 280k parameters (only about 1\% of the parameters), but significantly improves performance on unseen views. It further has significantly better runtime performance than spherical CNNs.

\begin{table}[h]
    \centering
    \caption{Comparison of the number of parameters in the 3 main models. Adding the geometric projection layer only adds 280k parameters, but greatly improves performance.}
    \label{tab:params}
    \begin{tabular}{l|cc}
    \toprule
    Model & \# params & $\Delta$ \\
    \midrule
    (2+1)D ResNet-50    & 21.3M & 0 \\
    (2+1)D ResNet-50 + Ours    & 21.5M & 280k  \\ 
    Spherical CNNs & 21.2M & -123k \\
    \bottomrule
    \end{tabular}
\end{table}

\section{Full Results}
The full numerical results from plots in the paper are provided here.

\begin{table}
    \centering
        \caption{How many cameras to use.}
    \label{tab:numcamera}
    \begin{tabular}{c|cc|cc}
    \toprule
    Method & \multicolumn{2}{c|}{MLB} & \multicolumn{2}{c}{TSH} \\
    & Seen & Unseen & Seen & Unseen \\
    \midrule
    Baseline & 55.6 & 30.2 & 49.8 & 34.2 \\
    1 Cam & 57.4 & 38.6 & 53.2 & 38.5 \\
    2 Cams & 58.1 & 41.8 & 53.9 & 39.1 \\
    4 Cams & 58.9 & 42.7 & 54.5 & 39.6 \\
    8 Cams & 58.7 & 42.7 & 54.5 & 39.4 \\
    \bottomrule
    \end{tabular}
\end{table}
\begin{table}
    \centering
        \caption{Where in network to add layer.}
    \label{tab:where}
    \begin{tabular}{c|cc|cc}
    \toprule
    Method & \multicolumn{2}{c|}{MLB} & \multicolumn{2}{c}{TSH} \\
    & Seen & Unseen & Seen & Unseen \\
    \midrule
    Block 1 & 57.8 & 42.1 & 54.3 & 39.2 \\
    Block 2 & 58.3 & 42.4 & 54.4 & 39.2  \\
    Block 3 & 58.9 & 42.7 & 54.5 & 39.6 \\
    Block 4 & 57.4 & 41.7 & 53.8 & 38.9 \\
    Block 5 & 57.1 & 40.9 & 53.3 & 37.7 \\
    \bottomrule
    \end{tabular}
\end{table}

\onecolumn
\section{PyTorch Implementation}
We provide the code here to implement the camera projection layer. 

\pythonexternal{camera_layer.py}

This layer can easily be inserted anywhere into a CNN. For example, assume the following code generates a ResNet. Then the camera transform is used as:

\begin{python}
class Net(nn.Module):
    def __init__(self, ...):
      self.layers = # ResNet Layers
      self.cam_props = CameraProps(channels)
      self.camera_proj = CameraProjection(num_cams)
    
    def forward(self, video):
      x = video
      for i,layer in enumerate(self.layers):
        x = layer(x)
        if i = apply_camera_layer_loc:
          rot, trans = self.cam_props(x)
          x = self.camera_proj(x, rot, trans)
      return x
\end{python}

\end{document}